\documentclass[
 reprint,
 amsmath,amssymb,
 aps, prl
]{revtex4-2}

\usepackage{graphicx}
\usepackage{dcolumn}
\usepackage{bm}
\usepackage{hyperref}

\begin{document}

\title{Robust Spectral Anomaly Detection in EELS Spectral Images via Three Dimensional Convolutional Variational Autoencoders}

\author{Seyfal Sultanov}
\author{James P. Buban}
\author{Robert F. Klie}
\email{rfklie@uic.edu}
\affiliation{Department of Computer Science, University of Illinois Chicago, Chicago, Illinois 60607, USA}
\affiliation{Department of Physics, University of Illinois Chicago, Chicago, Illinois 60607, USA}

\date{\today}

\begin{abstract}
We introduce a Three-Dimensional Convolutional Variational Autoencoder (3D-CVAE) for automated anomaly detection in Electron Energy Loss Spectroscopy Spectrum Imaging (EELS-SI) data. Our approach leverages the full three-dimensional structure of EELS-SI data to detect subtle spectral anomalies while preserving both spatial and spectral correlations across the datacube. By employing negative log-likelihood loss and training on bulk spectra, the model learns to reconstruct bulk features characteristic of the defect-free material. In exploring methods for anomaly detection, we evaluated both our 3D-CVAE approach and Principal Component Analysis (PCA), testing their performance using Fe \textit{L}-edge peak shifts designed to simulate material defects.
Our results show that 3D-CVAE achieves superior anomaly detection and maintains consistent performance across various shift magnitudes. The method demonstrates clear bimodal separation between normal and anomalous spectra, enabling reliable classification. Further analysis verifies that lower dimensional representations are robust to anomalies in the data. While performance advantages over PCA diminish with decreasing anomaly concentration, our method maintains high reconstruction quality even in challenging, noise-dominated spectral regions. This approach provides a robust framework for unsupervised  automated detection of spectral anomalies in EELS-SI data, particularly valuable for analyzing complex material systems.
\end{abstract}

\maketitle

\textit{\textbf{Introduction}}—High-resolution transmission electron microscopy has become the predominant approach for characterizing a wide range of materials, ranging from 2D materials \cite{RN1} to superconductors, \cite{RN2} semiconductors, \cite{RN3} and catalysts. \cite{RN4} A particularly powerful approach to materials characterization is the combination of scanning transmission electron microscopy (STEM) \cite{RN5} with electron energy-loss spectroscopy (EELS), \cite{RN6} which can measure the local density of states up to single atomic-column resolutions. \cite{RN7} This approach is often referred to as EELS spectrum imaging (EELS-SI), \cite{RN8} and the resulting 3-dimensional data cubes contain a detailed map of elemental composition, electronic structure, and bonding at the atomic scale, which is critical for understanding the fundamental properties of condensed matter systems. 

Core-loss EELS, which stems from the transition of highly-localized states, such as the \textit{1s} or \textit{2p} states into unoccupied orbitals above the Fermi-level $E_{F}$, often exhibit a detailed fine-structure of a particular edge, for example, the oxygen \textit{K}-edge or a transition metal \textit{L}-edge, near the edge onset, which reflects the density of unoccupied states near $E_{F}$. \cite{RN6} Subtle changes in this near-edge fine structure are due to changes in the local crystal structure, changes in orbital or spin ordering, valence state changes or the presence of defects/vacancies. These insights are invaluable for exploring phenomena, such as superconductivity, \cite{RN9} magnetism, \cite{RN10} and topological states of matter. \cite{RN11} Furthermore, atomic-column resolved EELS is particularly impactful in analyzing interfaces, grain boundaries, \cite{RN12} and low-dimensional materials, \cite{RN13} where local electronic and chemical environments dictate macroscopic material properties. By bridging the gap between atomic-scale phenomena and bulk material behavior, this technique plays a crucial role in advancing the design of quantum materials, catalysts, and energy devices.

Existing EELS-SI data analysis methods predominantly rely on manual inspection or dimensionality reduction techniques, such as Principal Component Analysis (PCA). While effective at noise reduction and extracting statistically significant features, PCA's linear nature limits its ability to capture physically significant, intricate spectral details. Its variance-based decomposition often relegates subtle spectral features to low-variance components, which are commonly discarded as noise. Furthermore, PCA, being constrained to linear combinations of input features, cannot accurately represent non-linear relationships in the data, potentially overlooking complex spectral patterns crucial for anomaly detection.

Machine learning (ML) has emerged as a significant tool across scientific disciplines, offering new approaches for analyzing complex datasets \cite{Mobarak2023,LeCun2015,Jordan2015}. In electron microscopy, conventional ML techniques have enhanced data analysis, enabling robust methods for denoising images and identifying atoms/patterns in STEM/STM/AFM data \cite{Lin2021,Hui2018,Somnath2018}. The increasing accessibility of high-performance computing has accelerated the adoption of more complex, data-intensive methods, particularly Deep Learning (DL) models like autoencoders, which have gained prominence in physics applications. Autoencoders, hourglass-shaped feed-forward neural networks, compress input data through an encoder, then reconstruct it from a low-dimensional representation, preserving salient features while finding a succinct data representation \cite{Hinton2006}.

Variational Autoencoders (VAEs) \cite{Kingma2013} combine variational inference and autoencoders to create deep generative models trainable in an unsupervised fashion. VAEs excel at learning compact, non-linear representations of high-dimensional data. They achieve this by regularizing the latent space so that nearby points encode semantically similar information. This regularization is accomplished by modeling the latent space as a product of Gaussian distributions and minimizing the Kullback-Leibler divergence between the estimated and true underlying distributions \cite{Kingma2013}. The Kullback-Leibler divergence is minimized when the estimated distribution matches the true underlying distribution, allowing the VAE to learn a smooth, continuous latent space that captures meaningful data features.

Unlike conventional autoencoders, which may learn discontinuous or arbitrary latent representations, VAEs' regularized latent space improves the quality of both learned features and the learned relationships between them. This characteristic makes VAEs more resistant to learning undesirable features such as noise signatures or subtle shifts in the training set - issues that often reduce the semantic meaning of latent encodings in conventional autoencoders - thereby enhancing VAEs' generalizability to new data.

In physics, VAEs have shown ability to learn physically relevant representations. For example, in molecular systems, VAEs have been applied to represent free energy surfaces (FES), enabling improved sampling of high-dimensional spaces and prediction of properties like isothermal compressibility or NMR spin-spin J couplings \cite{Carleo2019}. They have also been used for dimensionality reduction, such as identifying slowly varying collective variables in peptide folding, which is crucial for developing Markov state models of conformational changes \cite{Carleo2019}. 

In materials science, VAEs and related autoencoder architectures have been applied to extract meaningful features from various scientific images, including spatial-spectral characteristics from hyperspectral images using 3D convolutional autoencoders \cite{Mei2019} and structural patterns from STEM/STM images using shift-invariant VAEs \cite{Ziatdinov2023}. The latent space variables often correlate with key physical properties such as atomic positions, lattice periodicities, or electronic states, providing insights into the underlying physics. VAEs have demonstrated the capability to separate individual structural building blocks from relevant order parameter fields that change slowly on the length scale of the atomic lattice, enabling efficient exploration of complex configurational spaces \cite{Ziatdinov2023}.

Recent studies have applied ML to EELS data, creating models for predicting individual spectra from structural images based on the idea that local structures and functional phenomena are correlated through a small number of latent mechanisms \cite{Ziatdinov2022}. Denoising autoencoders have been explored as an alternative to PCA, matching and outperforming PCA reconstructions \cite{Pate2021}. However, most approaches have primarily addressed individual EELS spectra, leaving the full potential of 3D EELS-SI data unexplored.

VAEs have demonstrated effectiveness in anomaly detection across various domains. In civil engineering, they have been applied to detect temporal and spatial anomalies in dam monitoring data \cite{Shu2023}. In computer vision, VAEs have been used to detect and localize anomalous events in surveillance videos using only normal samples for training \cite{Fan2020}. In medical imaging, 3D VAEs have shown promise in detecting schizophrenia from brain MRI data \cite{Yamaguchi2021}.

Previous work focused on developing an approach using Convolutional VAEs (CVAEs) to detect and classify point defects and other structural anomalies in atomic-resolution STEM images \cite{Prifti2023,Ayyubi2024}. We successfully validated this method on STEM images of $SrTiO_{3}$ and more complex structures like $FePO_{4}$ and $CdTe$. In this approach, a CVAE trained solely on bulk crystal structure images learned the expected atomic positions and intensities. Anomalies were then identified by subtracting the input images from the CVAE’s reconstructions. 

The present work extends this concept to EELS-SI data, introducing a novel Three-Dimensional Convolutional Variational Autoencoder (3D-CVAE) for discovering intricate spectral anomalies. This unsupervised approach can learn complex, disentangled patterns in EELS-SI data while requiring relatively small training datasets compared to most supervised neural networks. Importantly, it can be implemented using computing resources widely available to researchers in the field, without requiring high performance supercomputers. 

\begin{figure*}
    \includegraphics[width=\textwidth]{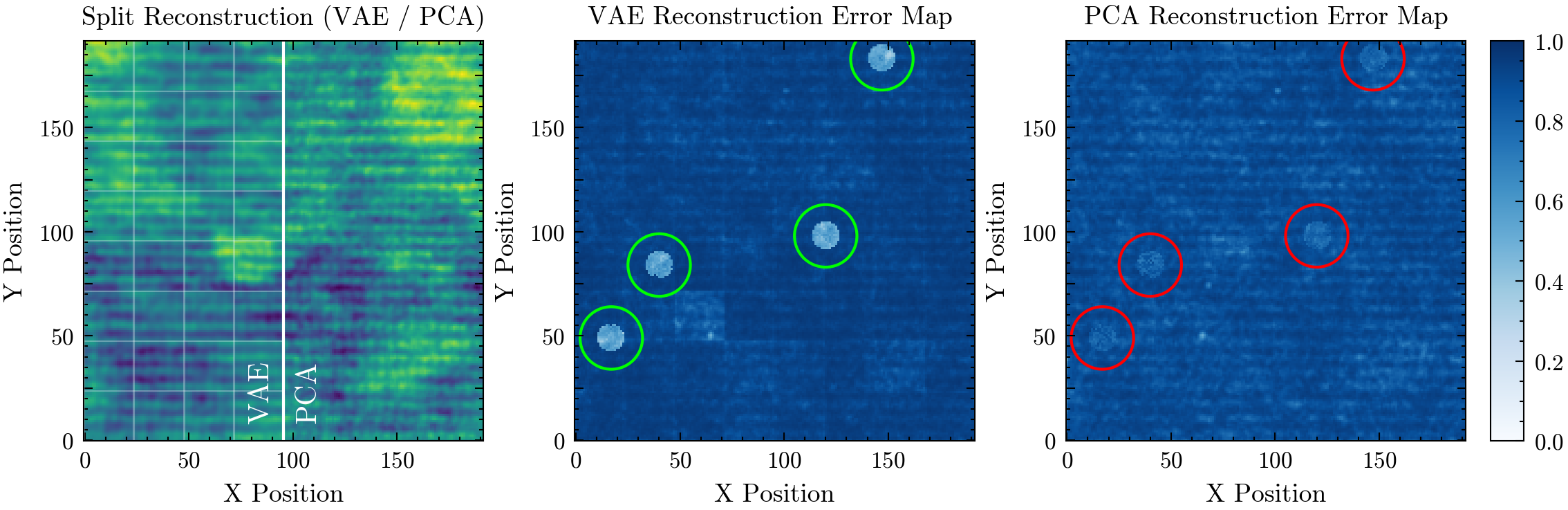}
    \caption{Comparison of VAE and PCA reconstructions and their anomaly detection performance. (a) Split visualization of the EELS-SI datacube reconstruction, with VAE (left) and PCA (4 components, right) results shown as z-direction intensity sums. (b) VAE reconstruction error heatmap showing Pearson Correlation Coefficients between original and reconstructed spectra. (c) Corresponding PCA reconstruction error heatmap. In both (b) and (c), green circles indicate successfully detected anomalies using Otsu's thresholding method, while red circles mark undetected anomalous regions. Lower PCC values (lighter colors) indicate greater deviation between original and reconstructed spectra.}
    \label{fig:3_heatmaps}
\end{figure*}

To enhance scalability and applicability, our model operates directly on EELS-SI data, eliminating the need for additional feature engineering or comprehensive prior knowledge of the material system. This element-agnostic approach allows the model to learn underlying spectral patterns for any element, given sufficient training examples.

We evaluate our method by training the 3D-CVAE on bulk EELS-SI data and testing its performance in reconstructing spectra with eight different types of artificially injected anomalies. Our results demonstrate that the 3D-CVAE-based method outperforms traditional PCA in both spectral reconstruction and anomaly detection.

For our experiments, we used EELS-SI spectra from epitaxial $BiFeO_{3}$ thin films grown on $SrTiO_{3}$

This approach successfully identifies subtle spectral changes associated with defect structures and interface phenomena, surpassing the capabilities of conventional analysis methods.

\textit{\textbf{Methods}}—The 3D-CVAE employs 3D convolutional layers to simultaneously capture spectral features and their spatial relationships within the EELS-SI datacube. Translational invariance is achieved through strided convolutions \cite{Lecun1998} across all three dimensions, ensuring consistent feature detection regardless of the exact position of spectral features. This architecture efficiently processes the full three-dimensional structure of the data while maintaining spatial relationships.

During training, the 3D-CVAE approximates the underlying data distribution by modeling it as a multivariate Gaussian distribution in a continuous latent space. In practice, the network learns to estimate the parameter space that generates this approximate distribution, with a Kullback-Leibler (KL) divergence term ensuring smoothness and preventing overfitting. The model encodes each spectrum as parameters (mean and variance) of this distribution in the latent space, where similar spectra cluster together based on their shared structural characteristics. During inference, when presented with an anomalous image, the VAE's encoder maps it into this learned latent space \cite{Kingma2013}. The subsequent reconstruction by the decoder is based on this mapping, effectively filtering out features that deviate from the learned data distribution. This process can be understood as a form of probabilistic dimensionality reduction followed by a generative reconstruction, where the model's learned prior acts as a constraint that guides the reconstruction towards the bulk structure of the training data. Consequently, the reconstructed image tends to exclude or attenuate elements that fall outside the learned distribution. The discrepancy between the original input and its reconstruction can then serve as a quantitative measure of anomaly, making VAEs an effective tool for both detecting and localizing anomalies in complex, high-dimensional data such as EELS SI datacube.

We present a novel DL method for EELS data by reformulating the reconstruction problem through Cross-Entropy Loss (CE). While existing DL approaches to spectral data typically employ Mean Squared Error (MSE) \cite{Pate2021} or Evidence Lower Bound (ELBO) \cite{Kingma2013} objectives that treat spectral intensities as continuous values, our formulation recognizes the discrete nature of electron energy loss events. Each spectrum in EELS represents a distribution of discrete electron counting events across energy channels. By utilizing Cross-Entropy Loss (CE) instead of MSE or ELBO, we treat each energy channel as a distinct class, where the normalized spectrum intensities represent the probabilities of electron energy loss events. This formulation aligns more closely with the probabilistic nature of the data and improves the model's ability to capture and reconstruct critical spectral features. The total loss function used for training combines the CE Loss term with a Kullback-Leibler (KL) divergence term, as follows:
\begin{align}
L_{\text{total}}(\mathbf{x}, \mathbf{\hat{x}}) = \underbrace{\sum_{i=1}^{N} L_{\text{CE}}(\mathbf{y}_i, \mathbf{\hat{y}}_i)}_{\text{Cross-Entropy Loss}} + \beta \cdot L_{\text{KL}}
\end{align}
Where $\mathbf{x}$ represents the input shard of the (SI) datacube, and $\mathbf{\hat{x}}$ represents the reconstructed shard produced by the decoder. The total number of spectra in a shard is denoted by $N$, which is obtained by flattening the spatial dimensions $(x,y)$ of the SI datacube. The parameter $\beta $ \cite{Higgins2017} is a weighting factor that controls the trade-off between the reconstruction accuracy (governed by the CE Loss) and the regularization of the latent space (enforced by the KL divergence term). The CE Loss quantifies the discrepancy between the original spectra and their reconstructions. For an individual spectrum, the CE Loss is defined as:
\begin{align}
L_{\text{CE}}(\mathbf{y}, \mathbf{\hat{y}}) = -\sum_{e=1}^{E} \left\{\ y_{e} \cdot \log[\text{softmax}(\hat{y}_{e})] \right \} 
\end{align} 
where $\mathbf{y} = \{y_{e}\}_{e=1}^{E}$ represents the normalized intensities of the original spectrum and $\mathbf{\hat{y}} = \{\hat{y}_{e}\}_{e=1}^{E}$ represents the reconstructed normalized intensities. The number of energy channels in each spectrum is denoted by $E$. The softmax function, $\text{softmax}(\hat{y}_{e})$, normalizes the reconstructed intensities to ensure that they are treated as probabilities, with values that sum to 1 across all energy channels. This formulation treats each energy channel as a distinct class, where the original normalized intensities $y_{e}$ represent the probability of observing an electron energy loss event in channel $e$. By optimizing this loss, the model reconstructs the spectra in a way that matches the probabilistic distribution of the original input data. In addition to the reconstruction loss, the KL divergence term regularizes the organization of the latent space, ensuring that it is smooth and aligned with a prior Gaussian distribution,. The KL divergence is defined as:
\begin{align}
L_{\text{KL}} = -\frac{1}{2} \sum_{j=1}^{J} \left[1 + \log(\sigma_{j}^2) - \mu_{j}^2 - \sigma_{j}^2\right]
\end{align}
Here, $J$ is the dimensionality of the latent space. The terms $\mu_{j}$ and $\sigma_{j}^2$ are the mean and variance of the approximate posterior distribution for the $j$-th latent dimension, respectively. This term encourages the latent representations to be close to the standard Gaussian prior, promoting a compact and well-organized latent space. The parameter $\beta$ in the total loss function governs the balance between the strength of this regularization and the fidelity of spectral reconstructions. Higher values of $\beta$  \cite{Higgins2017} enforce stricter regularization at the cost of reconstruction accuracy, while lower values prioritize precise reconstructions of the input spectra \cite{Kingma2013}. Through hyperparameter tuning, we determined $\beta = 1.2$ to provide the optimal balance between latent space organization and reconstruction quality for this specific dataset.

\textit{\textbf{Results}}—To validate our approach, we inject synthetic anomalies in the form of Fe \textit{L}-edge peak shift anomalies in spatially clustered patterns, simulating realistic defect structures. The Fe \textit{L}-edge was specifically chosen for this proof of concept due to its characteristically high Signal-to-Noise Ratio (SNR). 

\begin{figure}[b]
    \centering
    \includegraphics[width=1\linewidth]{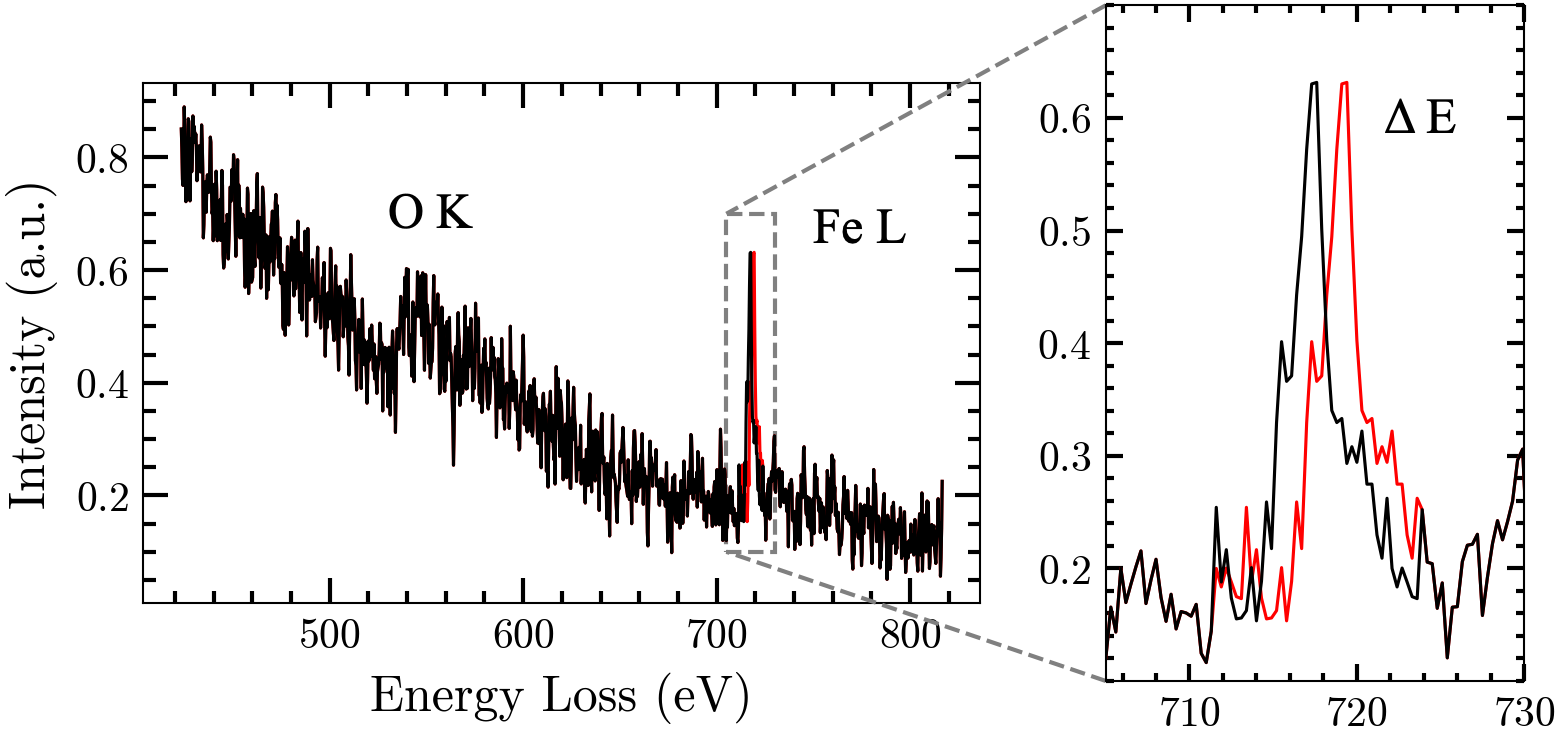}
    \caption{Example of an injected peak shift anomaly in EELS spectra. The original Fe L-edge spectrum (black) compared to an artificially introduced 2.5 eV peak shift (red segment)}
    \label{fig:anomaly_plot}
\end{figure}

The injected anomalies consist of 2.5 eV peak shifts, chosen to represent realistic defect-induced changes in electronic structure. Figure ~\ref{fig:anomaly_plot} demonstrates an example of such an anomaly, showing the original Fe L-edge spectrum (black) compared to the anomaly-injected spectrum (red), highlighting the characteristic peak shift our model aims to detect.
To evaluate detection performance, we compare our VAE-based approach against PCA reconstructions. The analysis pipeline processes the anomaly-injected datacube through both methods. For the VAE analysis, we segment the datacube into 24×24×L voxel blocks (where L represents the spectral dimension), process these through the network, and recombine them to preserve the original dimensions. For quantitative comparison, we calculate the Pearson Correlation Coefficient (PCC) between original and reconstructed spectra within the Fe L-edge energy window (690-730 eV) for both methods.
\begin{figure}[b]
    \centering
    \includegraphics[width=1\linewidth]{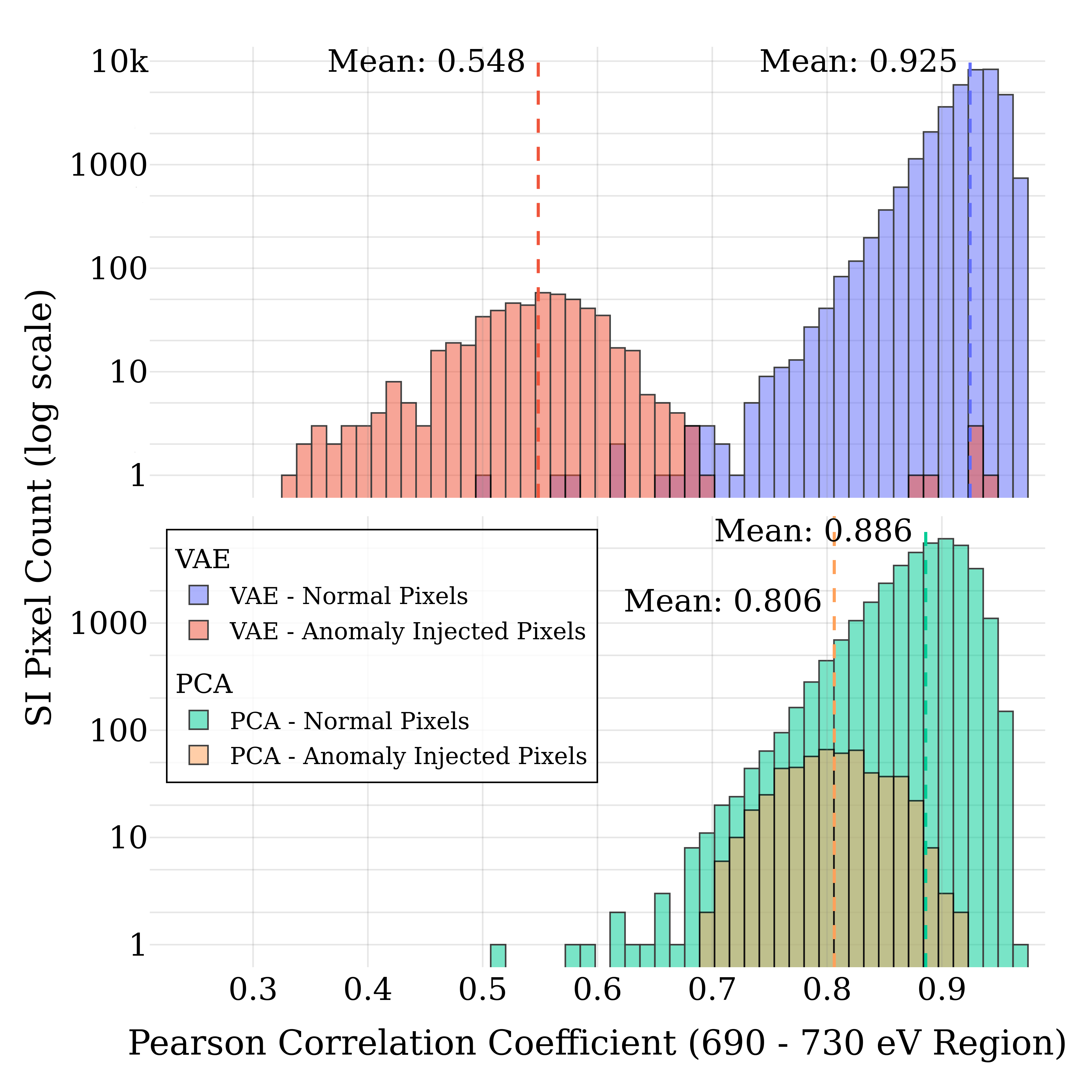}
    \caption{Distribution of pixels across Pearson Correlation Coefficient (PCC) values for VAE (top) and PCA with 4 components (bottom). Each histogram shows the distribution of normal (bulk) and anomalous pixels on a logarithmic scale. PCC values range from 0.2 to 1.0, where 1.0 indicates perfect correlation between original and reconstructed spectra.The VAE shows clear bimodal separation between normal and anomalous distributions, enabling reliable anomaly detection, while PCA distributions remain overlapped.}
    \label{fig:double_histogram}
\end{figure}
The PCC \cite{Pearson1895} metric measures the linear correlation between two variables, ranging from \(-1\) to \(1\), and is given by:
\begin{equation}
\rho_{x,y} = \frac{\sum_{e=1}^E (x_e - \bar{x})(y_e - \bar{y})}
{\sqrt{\sum_{e=1}^E (x_e - \bar{x})^2} \cdot \sqrt{\sum_{e=1}^E (y_e - \bar{y})^2}}
\end{equation}
In this context, the spectrum is treated as a pair of multivariate data vectors, \( \mathbf{x} = (x_1, x_2, \dots, x_E) \) and \( \mathbf{y} = (y_1, y_2, \dots, y_E) \), where \( E \) denotes the total number of energy channels. The variables \( x_e \) and \( y_e \) represent the intensities at the \( e \)-th energy channel for the respective spectra. The mean intensities of the spectra are given by \( \bar{x} = \frac{1}{E} \sum_{e=1}^E x_e \) and \( \bar{y} = \frac{1}{E} \sum_{e=1}^E y_e \), which capture the average intensity across all energy channels in each spectrum.

\begin{figure}[t]
    \centering
    \includegraphics[width=1\linewidth]{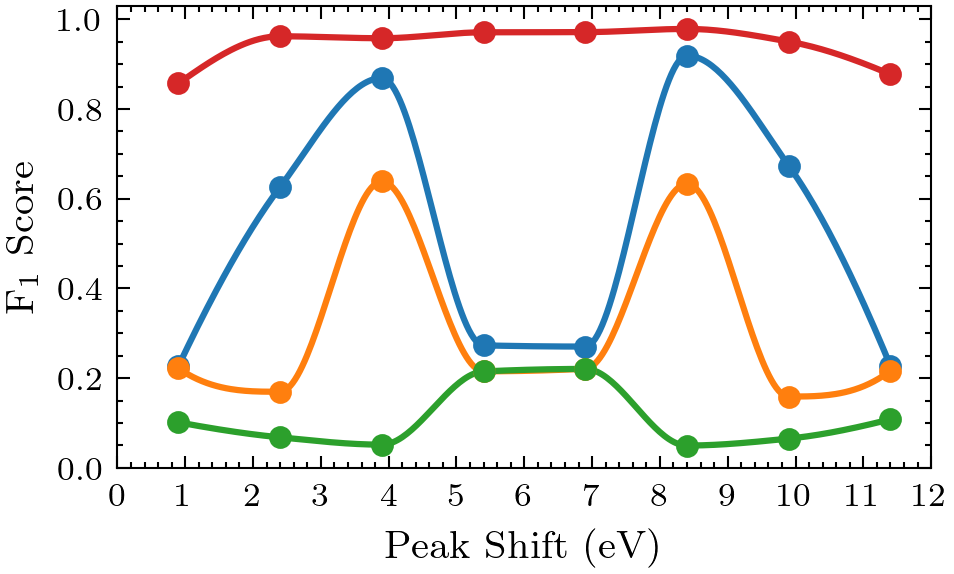}
    \caption{Performance comparison across different magnitudes of peak shifts. F1-scores for VAE (red) and PCA with 3 (blue), 4 (orange), and 5 (green) components. VAE maintains consistently high F1-scores across all shift magnitudes, while PCA exhibits periodic fluctuations in performance. PCA with 3 components shows the best performance among PCA variants, though its effectiveness varies with shift magnitude.}
    \label{fig:f1_score_plot}
\end{figure}

\begin{figure} [t]
    \centering
    \includegraphics[width=1\linewidth]{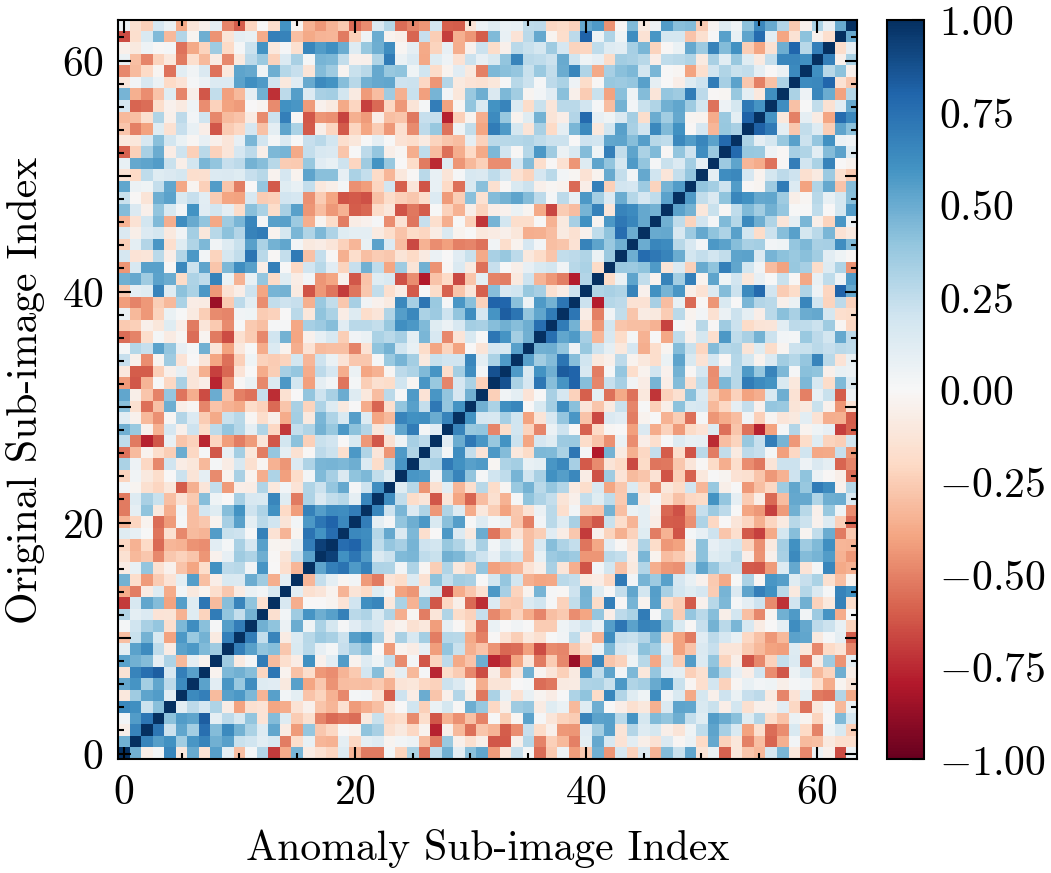}
    \caption{Visualization of latent space relationships through cosine similarity between 40-dimensional encodings of EELS-SI sub-image pairs. Each point compares an unmodified image (Y axis) with its anomaly-injected counterpart (X axis). The diagonal values close to 1 demonstrate that the encoder places paired images in nearly identical positions in the latent space, confirming that our model successfully recognizes anomalous spectra as variants of their normal counterparts.}
    \label{fig:latent_heatmap}
\end{figure}

Figure ~\ref{fig:3_heatmaps} provides a comprehensive visualization of both methods' performance highlighting the true positive areas detected by the VAE. To quantitatively identify anomalies, we analyze the distribution of Pearson Correlation Coefficient (PCC) scores across all pixels. As shown in Figure ~\ref{fig:3_heatmaps}, the VAE-generated error maps provide clearer visualization of localized errors, while the corresponding PCC distributions (Figure ~\ref{fig:double_histogram}) exhibit distinct bimodality, effectively separating normal and anomalous pixels. In contrast, while PCA-generated error maps show lower mean PCC values for anomalous regions, the distribution lacks clear separation between normal and anomalous populations, making reliable classification challenging.
For automated anomaly classification, we implement Otsu's method \cite{Otsu1979}, an algorithm that optimally separates the PCC histogram into two classes by maximizing between-class variance. To minimize false positives in anomaly-free data, we incorporate a unimodality check of the PCC distribution. In the specific case presented in Figures ~\ref{fig:3_heatmaps} and ~\ref{fig:double_histogram}, our VAE approach achieved high classification accuracy, with only 6 anomalous individual spectra misclassified as part of the bulk material structure out of 38,000 total spectra.
To verify the robustness of our method, we conducted a statistical analysis comparing our approach with PCA using 3, 4, and 5 components across various shift magnitudes (Figure ~\ref{fig:f1_score_plot}). Performance was evaluated using F1-scores \cite{Van1979, LeCun2015}, a harmonic mean of precision and recall that balances detection accuracy by accounting for both undetected anomalies (false negatives) and misclassified normal pixels (false positives). Our VAE approach demonstrates consistently high performance, maintaining robust F1-scores across different shift magnitudes with both high precision and recall. PCA performance shows a fundamental trade-off: using three components provides the best anomaly detection among PCA variants but exhibits periodic performance fluctuations based on shift-basis vector alignment. Adding more components improves spectral reconstruction fidelity but degrades anomaly detection capability, a limitation most apparent when examining subtle spectral features such as the O K edge.
Further analysis reveals that PCA performance is optimal when anomalies are small in number and sparsely distributed, while our VAE approach maintains consistent performance even beyond physically realistic anomaly concentrations. Although peak shifts beyond 4 eV exceed typical physical scenarios, we extended our analysis to larger shifts to demonstrate that PCA's trend of improving performance from 0 to 4 eV does not persist beyond this threshold and to show periodic nature of PCA performance.
To gain insight into the network's internal representations, we analyzed the latent space encodings of 64 pairs of EELS-SI sub-images, where each pair consisted of an unmodified datacube shard and its anomaly-injected counterpart. Analysis of cosine similarity between the 40-dimensional encodings (corresponding to our model's latent space dimensionality) reveals that the encoder consistently places paired images in close proximity within the latent space, as shown by the high correlation values along the diagonal in Figure ~\ref{fig:latent_heatmap}. This proximity is crucial for our anomaly detection approach, as it demonstrates that the encoder recognizes anomalous and normal spectra as fundamentally the same data point, leading to reconstructions that effectively filter out the anomalous features. This behavior confirms that our encoder successfully learns to represent the underlying normal spectral features while being robust to anomalous variations.

\textbf{Conclusion}—We have demonstrated a novel approach for automated anomaly detection in EELS-SI data using a 3D Convolutional Variational Autoencoder. Our method consistently outperforms traditional PCA-based approaches across various shift magnitudes while preserving spectral detail fidelity, though this performance gap narrows with decreasing anomaly concentration. Analysis of the latent space representations reveals that the model develops sophisticated encoding strategies that adapt to local spectral features, enabling robust anomaly detection without compromising reconstruction quality. While manual analysis confirms that our VAE approach maintains high reconstruction quality and feature preservation in lower SNR regions such as the O K edge, challenges remain regarding the development of quantitative metrics that can reliably assess reconstruction performance in these noise-dominated spectral regions. Future work will focus on establishing robust evaluation metrics that can better capture the demonstrated capabilities of our method, particularly for subtle spectral features where traditional pixel-by-pixel comparison metrics become unreliable due to noise. Additionally, recent advances in generative AI, particularly stable diffusion \cite{rombach2022} models, offer promising directions for EELS denoising and reconstruction. These models' ability to learn complex noise patterns and generate high-quality samples could potentially improve the recovery of fine spectral features in low-SNR regions.

\begin{acknowledgments}
\textbf{Acknowledgments}—Funding for this project was provided by the Alliance for Sustainable Energy, LLC, Managing and Operating Contractor for the National Renewable Energy Laboratory for the U.S. Department of Energy, and was supported in part by the U.S. Department of Energy's Office of Energy Efficiency and Renewable Energy (EERE) under the Solar Energy Technologies Office Award Number 37989. We thank M.E. Papka and the Electronic Visualization Laboratory for providing computational resources and hardware support, and R.A.W. Ayyubi for valuable discussions. The authors thank J.I. Idrobo for providing the sample EELS data and for the valuable discussions. 
\end{acknowledgments}


\nocite{*}

\bibliography{export}

\begin{thebibliography}{46}%
\makeatletter
\providecommand \@ifxundefined [1]{%
 \@ifx{#1\undefined}
}%
\providecommand \@ifnum [1]{%
 \ifnum #1\expandafter \@firstoftwo
 \else \expandafter \@secondoftwo
 \fi
}%
\providecommand \@ifx [1]{%
 \ifx #1\expandafter \@firstoftwo
 \else \expandafter \@secondoftwo
 \fi
}%
\providecommand \natexlab [1]{#1}%
\providecommand \enquote  [1]{``#1''}%
\providecommand \bibnamefont  [1]{#1}%
\providecommand \bibfnamefont [1]{#1}%
\providecommand \citenamefont [1]{#1}%
\providecommand \href@noop [0]{\@secondoftwo}%
\providecommand \href [0]{\begingroup \@sanitize@url \@href}%
\providecommand \@href[1]{\@@startlink{#1}\@@href}%
\providecommand \@@href[1]{\endgroup#1\@@endlink}%
\providecommand \@sanitize@url [0]{\catcode `\\12\catcode `\$12\catcode `\&12\catcode `\#12\catcode `\^12\catcode `\_12\catcode `\%12\relax}%
\providecommand \@@startlink[1]{}%
\providecommand \@@endlink[0]{}%
\providecommand \url  [0]{\begingroup\@sanitize@url \@url }%
\providecommand \@url [1]{\endgroup\@href {#1}{\urlprefix }}%
\providecommand \urlprefix  [0]{URL }%
\providecommand \Eprint [0]{\href }%
\providecommand \doibase [0]{http://dx.doi.org/}%
\providecommand \selectlanguage [0]{\@gobble}%
\providecommand \bibinfo  [0]{\@secondoftwo}%
\providecommand \bibfield  [0]{\@secondoftwo}%
\providecommand \translation [1]{[#1]}%
\providecommand \BibitemOpen [0]{}%
\providecommand \bibitemStop [0]{}%
\providecommand \bibitemNoStop [0]{.\EOS\space}%
\providecommand \EOS [0]{\spacefactor3000\relax}%
\providecommand \BibitemShut  [1]{\csname bibitem#1\endcsname}%
\let\auto@bib@innerbib\@empty
\bibitem [{\citenamefont {Song}\ \emph {et~al.}(2010)\citenamefont {Song}, \citenamefont {Ci}, \citenamefont {Lu}, \citenamefont {Sorokin}, \citenamefont {Jin}, \citenamefont {Ni}, \citenamefont {Kvashnin}, \citenamefont {Kvashnin}, \citenamefont {Lou}, \citenamefont {Yakobson},\ and\ \citenamefont {Ajayan}}]{RN1}%
  \BibitemOpen
  \bibfield  {author} {\bibinfo {author} {\bibfnamefont {L.}~\bibnamefont {Song}}, \bibinfo {author} {\bibfnamefont {L.~J.}\ \bibnamefont {Ci}}, \bibinfo {author} {\bibfnamefont {H.}~\bibnamefont {Lu}}, \bibinfo {author} {\bibfnamefont {P.~B.}\ \bibnamefont {Sorokin}}, \bibinfo {author} {\bibfnamefont {C.~H.}\ \bibnamefont {Jin}}, \bibinfo {author} {\bibfnamefont {J.}~\bibnamefont {Ni}}, \bibinfo {author} {\bibfnamefont {A.~G.}\ \bibnamefont {Kvashnin}}, \bibinfo {author} {\bibfnamefont {D.~G.}\ \bibnamefont {Kvashnin}}, \bibinfo {author} {\bibfnamefont {J.}~\bibnamefont {Lou}}, \bibinfo {author} {\bibfnamefont {B.~I.}\ \bibnamefont {Yakobson}}, \ and\ \bibinfo {author} {\bibfnamefont {P.~M.}\ \bibnamefont {Ajayan}},\ }\href {\doibase 10.1021/nl1022139} {\bibfield  {journal} {\bibinfo  {journal} {Nano Letters}\ }\textbf {\bibinfo {volume} {10}},\ \bibinfo {pages} {3209} (\bibinfo {year} {2010})}\BibitemShut {NoStop}%
\bibitem [{\citenamefont {Klie}\ \emph {et~al.}(2005)\citenamefont {Klie}, \citenamefont {Buban}, \citenamefont {Varela}, \citenamefont {Franceschetti}, \citenamefont {Jooss}, \citenamefont {Zhu}, \citenamefont {Browning}, \citenamefont {Pantelides},\ and\ \citenamefont {Pennycook}}]{RN2}%
  \BibitemOpen
  \bibfield  {author} {\bibinfo {author} {\bibfnamefont {R.~F.}\ \bibnamefont {Klie}}, \bibinfo {author} {\bibfnamefont {J.~P.}\ \bibnamefont {Buban}}, \bibinfo {author} {\bibfnamefont {M.}~\bibnamefont {Varela}}, \bibinfo {author} {\bibfnamefont {A.}~\bibnamefont {Franceschetti}}, \bibinfo {author} {\bibfnamefont {C.}~\bibnamefont {Jooss}}, \bibinfo {author} {\bibfnamefont {Y.}~\bibnamefont {Zhu}}, \bibinfo {author} {\bibfnamefont {N.~D.}\ \bibnamefont {Browning}}, \bibinfo {author} {\bibfnamefont {S.~T.}\ \bibnamefont {Pantelides}}, \ and\ \bibinfo {author} {\bibfnamefont {S.~J.}\ \bibnamefont {Pennycook}},\ }\href {<Go to ISI>://000229337800050} {\bibfield  {journal} {\bibinfo  {journal} {Nature}\ }\textbf {\bibinfo {volume} {435}},\ \bibinfo {pages} {475} (\bibinfo {year} {2005})}\BibitemShut {NoStop}%
\bibitem [{\citenamefont {Voyles}\ \emph {et~al.}(2002)\citenamefont {Voyles}, \citenamefont {Muller}, \citenamefont {Grazul}, \citenamefont {Citrin},\ and\ \citenamefont {Gossmann}}]{RN3}%
  \BibitemOpen
  \bibfield  {author} {\bibinfo {author} {\bibfnamefont {P.~M.}\ \bibnamefont {Voyles}}, \bibinfo {author} {\bibfnamefont {D.~A.}\ \bibnamefont {Muller}}, \bibinfo {author} {\bibfnamefont {J.~L.}\ \bibnamefont {Grazul}}, \bibinfo {author} {\bibfnamefont {P.~H.}\ \bibnamefont {Citrin}}, \ and\ \bibinfo {author} {\bibfnamefont {H.~J.~L.}\ \bibnamefont {Gossmann}},\ }\href {<Go to ISI>://000175163800045} {\bibfield  {journal} {\bibinfo  {journal} {Nature}\ }\textbf {\bibinfo {volume} {416}},\ \bibinfo {pages} {826} (\bibinfo {year} {2002})}\BibitemShut {NoStop}%
\bibitem [{\citenamefont {Sun}\ \emph {et~al.}(2002)\citenamefont {Sun}, \citenamefont {Liu}, \citenamefont {Nag},\ and\ \citenamefont {Browning}}]{RN4}%
  \BibitemOpen
  \bibfield  {author} {\bibinfo {author} {\bibfnamefont {K.}~\bibnamefont {Sun}}, \bibinfo {author} {\bibfnamefont {J.}~\bibnamefont {Liu}}, \bibinfo {author} {\bibfnamefont {N.}~\bibnamefont {Nag}}, \ and\ \bibinfo {author} {\bibfnamefont {N.~D.}\ \bibnamefont {Browning}},\ }\href {<Go to ISI>://000179774300008} {\bibfield  {journal} {\bibinfo  {journal} {Catalysis Letters}\ }\textbf {\bibinfo {volume} {84}},\ \bibinfo {pages} {193} (\bibinfo {year} {2002})}\BibitemShut {NoStop}%
\bibitem [{\citenamefont {Pennycook}\ and\ \citenamefont {Boatner}(1988)}]{RN5}%
  \BibitemOpen
  \bibfield  {author} {\bibinfo {author} {\bibfnamefont {S.~J.}\ \bibnamefont {Pennycook}}\ and\ \bibinfo {author} {\bibfnamefont {L.~A.}\ \bibnamefont {Boatner}},\ }\href {<Go to ISI>://A1988R248000066} {\bibfield  {journal} {\bibinfo  {journal} {Nature}\ }\textbf {\bibinfo {volume} {336}},\ \bibinfo {pages} {565} (\bibinfo {year} {1988})}\BibitemShut {NoStop}%
\bibitem [{\citenamefont {Egerton}(2011)}]{RN6}%
  \BibitemOpen
  \bibfield  {author} {\bibinfo {author} {\bibfnamefont {R.}~\bibnamefont {Egerton}},\ }\href@noop {} {\emph {\bibinfo {title} {Electron Energy Loss Spectroscopy in the Electron Microscope}}},\ \bibinfo {edition} {2nd}\ ed.\ (\bibinfo  {publisher} {Springer Science \& Business Media},\ \bibinfo {address} {New York},\ \bibinfo {year} {2011})\BibitemShut {NoStop}%
\bibitem [{\citenamefont {Varela}\ \emph {et~al.}(2004)\citenamefont {Varela}, \citenamefont {Findlay}, \citenamefont {Lupini}, \citenamefont {Christen}, \citenamefont {Borisevich}, \citenamefont {Dellby}, \citenamefont {Krivanek}, \citenamefont {Nellist}, \citenamefont {Oxley}, \citenamefont {Allen},\ and\ \citenamefont {Pennycook}}]{RN7}%
  \BibitemOpen
  \bibfield  {author} {\bibinfo {author} {\bibfnamefont {M.}~\bibnamefont {Varela}}, \bibinfo {author} {\bibfnamefont {S.~D.}\ \bibnamefont {Findlay}}, \bibinfo {author} {\bibfnamefont {A.~R.}\ \bibnamefont {Lupini}}, \bibinfo {author} {\bibfnamefont {H.~M.}\ \bibnamefont {Christen}}, \bibinfo {author} {\bibfnamefont {A.~Y.}\ \bibnamefont {Borisevich}}, \bibinfo {author} {\bibfnamefont {N.}~\bibnamefont {Dellby}}, \bibinfo {author} {\bibfnamefont {O.~L.}\ \bibnamefont {Krivanek}}, \bibinfo {author} {\bibfnamefont {P.~D.}\ \bibnamefont {Nellist}}, \bibinfo {author} {\bibfnamefont {M.~P.}\ \bibnamefont {Oxley}}, \bibinfo {author} {\bibfnamefont {L.~J.}\ \bibnamefont {Allen}}, \ and\ \bibinfo {author} {\bibfnamefont {S.~J.}\ \bibnamefont {Pennycook}},\ }\href {<Go to ISI>://000220055400030} {\bibfield  {journal} {\bibinfo  {journal} {Physical Review Letters}\ }\textbf {\bibinfo {volume} {92}},\ \bibinfo {pages} {095502} (\bibinfo {year} {2004})}\BibitemShut {NoStop}%
\bibitem [{\citenamefont {Jeanguillaume}\ and\ \citenamefont {Colliex}(1989)}]{RN8}%
  \BibitemOpen
  \bibfield  {author} {\bibinfo {author} {\bibfnamefont {C.}~\bibnamefont {Jeanguillaume}}\ and\ \bibinfo {author} {\bibfnamefont {C.}~\bibnamefont {Colliex}},\ }\href {\doibase 10.1016/0304-3991(89)90304-5} {\bibfield  {journal} {\bibinfo  {journal} {Ultramicroscopy}\ }\textbf {\bibinfo {volume} {28}},\ \bibinfo {pages} {252} (\bibinfo {year} {1989})}\BibitemShut {NoStop}%
\bibitem [{\citenamefont {Browning}\ \emph {et~al.}(1993)\citenamefont {Browning}, \citenamefont {Chisholm}, \citenamefont {Pennycook}, \citenamefont {Norton},\ and\ \citenamefont {Lowndes}}]{RN9}%
  \BibitemOpen
  \bibfield  {author} {\bibinfo {author} {\bibfnamefont {N.~D.}\ \bibnamefont {Browning}}, \bibinfo {author} {\bibfnamefont {M.~F.}\ \bibnamefont {Chisholm}}, \bibinfo {author} {\bibfnamefont {S.~J.}\ \bibnamefont {Pennycook}}, \bibinfo {author} {\bibfnamefont {D.~P.}\ \bibnamefont {Norton}}, \ and\ \bibinfo {author} {\bibfnamefont {D.~H.}\ \bibnamefont {Lowndes}},\ }\href {<Go to ISI>://A1993LL33900024} {\bibfield  {journal} {\bibinfo  {journal} {Physica C}\ }\textbf {\bibinfo {volume} {212}},\ \bibinfo {pages} {185} (\bibinfo {year} {1993})}\BibitemShut {NoStop}%
\bibitem [{\citenamefont {Klie}\ \emph {et~al.}(2007)\citenamefont {Klie}, \citenamefont {Zheng}, \citenamefont {Zhu}, \citenamefont {Varela}, \citenamefont {Wu},\ and\ \citenamefont {Leighton}}]{RN10}%
  \BibitemOpen
  \bibfield  {author} {\bibinfo {author} {\bibfnamefont {R.~F.}\ \bibnamefont {Klie}}, \bibinfo {author} {\bibfnamefont {J.~C.}\ \bibnamefont {Zheng}}, \bibinfo {author} {\bibfnamefont {Y.}~\bibnamefont {Zhu}}, \bibinfo {author} {\bibfnamefont {M.}~\bibnamefont {Varela}}, \bibinfo {author} {\bibfnamefont {J.}~\bibnamefont {Wu}}, \ and\ \bibinfo {author} {\bibfnamefont {C.}~\bibnamefont {Leighton}},\ }\href {http://link.aps.org/abstract/PRL/v99/e047203} {\bibfield  {journal} {\bibinfo  {journal} {Physical Review Letters}\ }\textbf {\bibinfo {volume} {99}},\ \bibinfo {pages} {047203} (\bibinfo {year} {2007})}\BibitemShut {NoStop}%
\bibitem [{\citenamefont {Li}\ \emph {et~al.}(2015)\citenamefont {Li}, \citenamefont {Chang}, \citenamefont {Wu}, \citenamefont {Tao}, \citenamefont {Zhao}, \citenamefont {Chan}, \citenamefont {Moodera}, \citenamefont {Li},\ and\ \citenamefont {Zhu}}]{RN11}%
  \BibitemOpen
  \bibfield  {author} {\bibinfo {author} {\bibfnamefont {M.~D.}\ \bibnamefont {Li}}, \bibinfo {author} {\bibfnamefont {C.~Z.}\ \bibnamefont {Chang}}, \bibinfo {author} {\bibfnamefont {L.~J.}\ \bibnamefont {Wu}}, \bibinfo {author} {\bibfnamefont {J.}~\bibnamefont {Tao}}, \bibinfo {author} {\bibfnamefont {W.~W.}\ \bibnamefont {Zhao}}, \bibinfo {author} {\bibfnamefont {M.~H.~W.}\ \bibnamefont {Chan}}, \bibinfo {author} {\bibfnamefont {J.~S.}\ \bibnamefont {Moodera}}, \bibinfo {author} {\bibfnamefont {J.}~\bibnamefont {Li}}, \ and\ \bibinfo {author} {\bibfnamefont {Y.~M.}\ \bibnamefont {Zhu}},\ }\href {\doibase 10.1103/PhysRevLett.114.146802} {\bibfield  {journal} {\bibinfo  {journal} {Physical Review Letters}\ }\textbf {\bibinfo {volume} {114}} (\bibinfo {year} {2015}),\ 10.1103/PhysRevLett.114.146802}\BibitemShut {NoStop}%
\bibitem [{\citenamefont {Klie}\ and\ \citenamefont {Browning}(2000)}]{RN12}%
  \BibitemOpen
  \bibfield  {author} {\bibinfo {author} {\bibfnamefont {R.~F.}\ \bibnamefont {Klie}}\ and\ \bibinfo {author} {\bibfnamefont {N.~D.}\ \bibnamefont {Browning}},\ }\href {<Go to ISI>://000165584700019} {\bibfield  {journal} {\bibinfo  {journal} {Applied Physics Letters}\ }\textbf {\bibinfo {volume} {77}},\ \bibinfo {pages} {3737} (\bibinfo {year} {2000})}\BibitemShut {NoStop}%
\bibitem [{\citenamefont {Lagunas}\ \emph {et~al.}(2024)\citenamefont {Lagunas}, \citenamefont {Bugallo}, \citenamefont {Karimi}, \citenamefont {Yang}, \citenamefont {Badr}, \citenamefont {Cope}, \citenamefont {Ferral}, \citenamefont {Barsoum}, \citenamefont {Hu},\ and\ \citenamefont {Klie}}]{RN13}%
  \BibitemOpen
  \bibfield  {author} {\bibinfo {author} {\bibfnamefont {F.}~\bibnamefont {Lagunas}}, \bibinfo {author} {\bibfnamefont {D.}~\bibnamefont {Bugallo}}, \bibinfo {author} {\bibfnamefont {F.}~\bibnamefont {Karimi}}, \bibinfo {author} {\bibfnamefont {Y.~J.}\ \bibnamefont {Yang}}, \bibinfo {author} {\bibfnamefont {H.~O.}\ \bibnamefont {Badr}}, \bibinfo {author} {\bibfnamefont {J.~H.}\ \bibnamefont {Cope}}, \bibinfo {author} {\bibfnamefont {E.}~\bibnamefont {Ferral}}, \bibinfo {author} {\bibfnamefont {M.~W.}\ \bibnamefont {Barsoum}}, \bibinfo {author} {\bibfnamefont {Y.~J.}\ \bibnamefont {Hu}}, \ and\ \bibinfo {author} {\bibfnamefont {R.~F.}\ \bibnamefont {Klie}},\ }\href {\doibase 10.1021/acs.chemmater.3c02773} {\bibfield  {journal} {\bibinfo  {journal} {Chemistry of Materials}\ }\textbf {\bibinfo {volume} {36}},\ \bibinfo {pages} {2743} (\bibinfo {year} {2024})}\BibitemShut {NoStop}%
\bibitem [{\citenamefont {Mobarak}\ \emph {et~al.}(2023)\citenamefont {Mobarak}, \citenamefont {Mimona}, \citenamefont {Islam}, \citenamefont {Hossain}, \citenamefont {Zohura}, \citenamefont {Imtiaz},\ and\ \citenamefont {Rimon}}]{Mobarak2023}%
  \BibitemOpen
  \bibfield  {author} {\bibinfo {author} {\bibfnamefont {M.~H.}\ \bibnamefont {Mobarak}}, \bibinfo {author} {\bibfnamefont {M.~A.}\ \bibnamefont {Mimona}}, \bibinfo {author} {\bibfnamefont {M.~A.}\ \bibnamefont {Islam}}, \bibinfo {author} {\bibfnamefont {N.}~\bibnamefont {Hossain}}, \bibinfo {author} {\bibfnamefont {F.~T.}\ \bibnamefont {Zohura}}, \bibinfo {author} {\bibfnamefont {I.}~\bibnamefont {Imtiaz}}, \ and\ \bibinfo {author} {\bibfnamefont {M.~I.~H.}\ \bibnamefont {Rimon}},\ }\href {\doibase 10.1016/j.apsadv.2023.100523} {\bibfield  {journal} {\bibinfo  {journal} {Applied Surface Science Advances}\ }\textbf {\bibinfo {volume} {18}},\ \bibinfo {pages} {100523} (\bibinfo {year} {2023})}\BibitemShut {NoStop}%
\bibitem [{\citenamefont {LeCun}\ \emph {et~al.}(2015)\citenamefont {LeCun}, \citenamefont {Bengio},\ and\ \citenamefont {Hinton}}]{LeCun2015}%
  \BibitemOpen
  \bibfield  {author} {\bibinfo {author} {\bibfnamefont {Y.}~\bibnamefont {LeCun}}, \bibinfo {author} {\bibfnamefont {Y.}~\bibnamefont {Bengio}}, \ and\ \bibinfo {author} {\bibfnamefont {G.}~\bibnamefont {Hinton}},\ }\href {\doibase 10.1038/nature14539} {\bibfield  {journal} {\bibinfo  {journal} {Nature}\ }\textbf {\bibinfo {volume} {521}},\ \bibinfo {pages} {436} (\bibinfo {year} {2015})}\BibitemShut {NoStop}%
\bibitem [{\citenamefont {Jordan}\ and\ \citenamefont {Mitchell}(2015)}]{Jordan2015}%
  \BibitemOpen
  \bibfield  {author} {\bibinfo {author} {\bibfnamefont {M.~I.}\ \bibnamefont {Jordan}}\ and\ \bibinfo {author} {\bibfnamefont {T.~M.}\ \bibnamefont {Mitchell}},\ }\href {\doibase 10.1126/science.aaa8415} {\bibfield  {journal} {\bibinfo  {journal} {Science}\ }\textbf {\bibinfo {volume} {349}},\ \bibinfo {pages} {255} (\bibinfo {year} {2015})}\BibitemShut {NoStop}%
\bibitem [{\citenamefont {Lin}\ \emph {et~al.}(2021)\citenamefont {Lin}, \citenamefont {Zhang}, \citenamefont {Wang}, \citenamefont {Yang},\ and\ \citenamefont {Xin}}]{Lin2021}%
  \BibitemOpen
  \bibfield  {author} {\bibinfo {author} {\bibfnamefont {R.}~\bibnamefont {Lin}}, \bibinfo {author} {\bibfnamefont {R.}~\bibnamefont {Zhang}}, \bibinfo {author} {\bibfnamefont {C.}~\bibnamefont {Wang}}, \bibinfo {author} {\bibfnamefont {X.~Q.}\ \bibnamefont {Yang}}, \ and\ \bibinfo {author} {\bibfnamefont {H.~L.}\ \bibnamefont {Xin}},\ }\href {\doibase 10.1038/s41598-021-84499-w} {\bibfield  {journal} {\bibinfo  {journal} {Scientific Reports}\ }\textbf {\bibinfo {volume} {11}} (\bibinfo {year} {2021}),\ 10.1038/s41598-021-84499-w}\BibitemShut {NoStop}%
\bibitem [{\citenamefont {Hui}\ and\ \citenamefont {Liu}(2018)}]{Hui2018}%
  \BibitemOpen
  \bibfield  {author} {\bibinfo {author} {\bibfnamefont {Y.}~\bibnamefont {Hui}}\ and\ \bibinfo {author} {\bibfnamefont {Y.}~\bibnamefont {Liu}},\ }\href {https://arxiv.org/abs/1809.05076} {\enquote {\bibinfo {title} {Computer vision-aided atom tracking in stem imaging},}\ } (\bibinfo {year} {2018}),\ \Eprint {http://arxiv.org/abs/1809.05076}{arXiv:1809.05076 [cs.CV]}\BibitemShut {NoStop}%
\bibitem [{\citenamefont {Somnath}\ \emph {et~al.}(2018)\citenamefont {Somnath}, \citenamefont {Smith}, \citenamefont {Kalinin}, \citenamefont {Chi}, \citenamefont {Borisevich}, \citenamefont {Cross}, \citenamefont {Duscher},\ and\ \citenamefont {Jesse}}]{Somnath2018}%
  \BibitemOpen
  \bibfield  {author} {\bibinfo {author} {\bibfnamefont {S.}~\bibnamefont {Somnath}}, \bibinfo {author} {\bibfnamefont {C.~R.}\ \bibnamefont {Smith}}, \bibinfo {author} {\bibfnamefont {S.~V.}\ \bibnamefont {Kalinin}}, \bibinfo {author} {\bibfnamefont {M.}~\bibnamefont {Chi}}, \bibinfo {author} {\bibfnamefont {A.}~\bibnamefont {Borisevich}}, \bibinfo {author} {\bibfnamefont {N.}~\bibnamefont {Cross}}, \bibinfo {author} {\bibfnamefont {G.}~\bibnamefont {Duscher}}, \ and\ \bibinfo {author} {\bibfnamefont {S.}~\bibnamefont {Jesse}},\ }\href {\doibase 10.1186/s40679-018-0052-y} {\bibfield  {journal} {\bibinfo  {journal} {Advanced Structural and Chemical Imaging}\ }\textbf {\bibinfo {volume} {4}} (\bibinfo {year} {2018}),\ 10.1186/s40679-018-0052-y}\BibitemShut {NoStop}%
\bibitem [{\citenamefont {Hinton}\ and\ \citenamefont {Salakhutdinov}(2006)}]{Hinton2006}%
  \BibitemOpen
  \bibfield  {author} {\bibinfo {author} {\bibfnamefont {G.~E.}\ \bibnamefont {Hinton}}\ and\ \bibinfo {author} {\bibfnamefont {R.~R.}\ \bibnamefont {Salakhutdinov}},\ }\href {\doibase 10.1126/science.1127647} {\bibfield  {journal} {\bibinfo  {journal} {Science}\ }\textbf {\bibinfo {volume} {313}},\ \bibinfo {pages} {504} (\bibinfo {year} {2006})}\BibitemShut {NoStop}%
\bibitem [{\citenamefont {Kingma}\ and\ \citenamefont {Welling}(2022)}]{Kingma2013}%
  \BibitemOpen
  \bibfield  {author} {\bibinfo {author} {\bibfnamefont {D.~P.}\ \bibnamefont {Kingma}}\ and\ \bibinfo {author} {\bibfnamefont {M.}~\bibnamefont {Welling}},\ }\href {https://arxiv.org/abs/1312.6114} {\enquote {\bibinfo {title} {Auto-encoding variational bayes},}\ } (\bibinfo {year} {2022}),\ \Eprint {http://arxiv.org/abs/1312.6114}{arXiv:1312.6114 [stat.ML]}\BibitemShut {NoStop}%
\bibitem [{\citenamefont {Carleo}\ \emph {et~al.}(2019)\citenamefont {Carleo}, \citenamefont {Cirac}, \citenamefont {Cranmer}, \citenamefont {Daudet}, \citenamefont {Schuld}, \citenamefont {Tishby}, \citenamefont {Vogt-Maranto},\ and\ \citenamefont {Zdeborová}}]{Carleo2019}%
  \BibitemOpen
  \bibfield  {author} {\bibinfo {author} {\bibfnamefont {G.}~\bibnamefont {Carleo}}, \bibinfo {author} {\bibfnamefont {I.}~\bibnamefont {Cirac}}, \bibinfo {author} {\bibfnamefont {K.}~\bibnamefont {Cranmer}}, \bibinfo {author} {\bibfnamefont {L.}~\bibnamefont {Daudet}}, \bibinfo {author} {\bibfnamefont {M.}~\bibnamefont {Schuld}}, \bibinfo {author} {\bibfnamefont {N.}~\bibnamefont {Tishby}}, \bibinfo {author} {\bibfnamefont {L.}~\bibnamefont {Vogt-Maranto}}, \ and\ \bibinfo {author} {\bibfnamefont {L.}~\bibnamefont {Zdeborová}},\ }\href {\doibase 10.1103/RevModPhys.91.045002} {\bibfield  {journal} {\bibinfo  {journal} {Reviews of Modern Physics}\ }\textbf {\bibinfo {volume} {91}},\ \bibinfo {pages} {045002} (\bibinfo {year} {2019})}\BibitemShut {NoStop}%
\bibitem [{\citenamefont {Mei}\ \emph {et~al.}(2019)\citenamefont {Mei}, \citenamefont {Ji}, \citenamefont {Geng}, \citenamefont {Zhang}, \citenamefont {Li},\ and\ \citenamefont {Du}}]{Mei2019}%
  \BibitemOpen
  \bibfield  {author} {\bibinfo {author} {\bibfnamefont {S.}~\bibnamefont {Mei}}, \bibinfo {author} {\bibfnamefont {J.}~\bibnamefont {Ji}}, \bibinfo {author} {\bibfnamefont {Y.}~\bibnamefont {Geng}}, \bibinfo {author} {\bibfnamefont {Z.}~\bibnamefont {Zhang}}, \bibinfo {author} {\bibfnamefont {X.}~\bibnamefont {Li}}, \ and\ \bibinfo {author} {\bibfnamefont {Q.}~\bibnamefont {Du}},\ }\href {\doibase 10.1109/TGRS.2019.2908756} {\bibfield  {journal} {\bibinfo  {journal} {IEEE Transactions on Geoscience and Remote Sensing}\ }\textbf {\bibinfo {volume} {57}},\ \bibinfo {pages} {6808} (\bibinfo {year} {2019})}\BibitemShut {NoStop}%
\bibitem [{\citenamefont {Ziatdinov}\ \emph {et~al.}(2023)\citenamefont {Ziatdinov}, \citenamefont {Wong},\ and\ \citenamefont {Kalinin}}]{Ziatdinov2023}%
  \BibitemOpen
  \bibfield  {author} {\bibinfo {author} {\bibfnamefont {M.}~\bibnamefont {Ziatdinov}}, \bibinfo {author} {\bibfnamefont {C.~Y.~T.}\ \bibnamefont {Wong}}, \ and\ \bibinfo {author} {\bibfnamefont {S.~V.}\ \bibnamefont {Kalinin}},\ }\href {\doibase 10.1088/2632-2153/ad073b} {\bibfield  {journal} {\bibinfo  {journal} {Machine Learning: Science and Technology}\ }\textbf {\bibinfo {volume} {4}} (\bibinfo {year} {2023}),\ 10.1088/2632-2153/ad073b}\BibitemShut {NoStop}%
\bibitem [{\citenamefont {Ziatdinov}\ \emph {et~al.}(2022)\citenamefont {Ziatdinov}, \citenamefont {Ghosh}, \citenamefont {Wong},\ and\ \citenamefont {Kalinin}}]{Ziatdinov2022}%
  \BibitemOpen
  \bibfield  {author} {\bibinfo {author} {\bibfnamefont {M.}~\bibnamefont {Ziatdinov}}, \bibinfo {author} {\bibfnamefont {A.}~\bibnamefont {Ghosh}}, \bibinfo {author} {\bibfnamefont {T.}~\bibnamefont {Wong}}, \ and\ \bibinfo {author} {\bibfnamefont {S.~V.}\ \bibnamefont {Kalinin}},\ }\href {\doibase 10.1038/s42256-022-00555-8} {\bibfield  {journal} {\bibinfo  {journal} {Nature Machine Intelligence}\ }\textbf {\bibinfo {volume} {4}},\ \bibinfo {pages} {1101} (\bibinfo {year} {2022})}\BibitemShut {NoStop}%
\bibitem [{\citenamefont {Pate}\ \emph {et~al.}(2021)\citenamefont {Pate}, \citenamefont {Hart},\ and\ \citenamefont {Taheri}}]{Pate2021}%
  \BibitemOpen
  \bibfield  {author} {\bibinfo {author} {\bibfnamefont {C.~M.}\ \bibnamefont {Pate}}, \bibinfo {author} {\bibfnamefont {J.~L.}\ \bibnamefont {Hart}}, \ and\ \bibinfo {author} {\bibfnamefont {M.~L.}\ \bibnamefont {Taheri}},\ }\href {\doibase 10.1038/s41598-021-97668-8} {\bibfield  {journal} {\bibinfo  {journal} {Scientific Reports}\ }\textbf {\bibinfo {volume} {11}},\ \bibinfo {pages} {19515} (\bibinfo {year} {2021})}\BibitemShut {NoStop}%
\bibitem [{\citenamefont {Shu}\ \emph {et~al.}(2023)\citenamefont {Shu}, \citenamefont {Bao}, \citenamefont {Zhou}, \citenamefont {Xu}, \citenamefont {Li},\ and\ \citenamefont {Zhang}}]{Shu2023}%
  \BibitemOpen
  \bibfield  {author} {\bibinfo {author} {\bibfnamefont {X.}~\bibnamefont {Shu}}, \bibinfo {author} {\bibfnamefont {T.}~\bibnamefont {Bao}}, \bibinfo {author} {\bibfnamefont {Y.}~\bibnamefont {Zhou}}, \bibinfo {author} {\bibfnamefont {R.}~\bibnamefont {Xu}}, \bibinfo {author} {\bibfnamefont {Y.}~\bibnamefont {Li}}, \ and\ \bibinfo {author} {\bibfnamefont {K.}~\bibnamefont {Zhang}},\ }\href {\doibase 10.1177/14759217211073301} {\bibfield  {journal} {\bibinfo  {journal} {Structural Health Monitoring}\ }\textbf {\bibinfo {volume} {22}},\ \bibinfo {pages} {39} (\bibinfo {year} {2023})}\BibitemShut {NoStop}%
\bibitem [{\citenamefont {Fan}\ \emph {et~al.}(2020)\citenamefont {Fan}, \citenamefont {Wen}, \citenamefont {Li}, \citenamefont {Qiu}, \citenamefont {Levine},\ and\ \citenamefont {Xiao}}]{Fan2020}%
  \BibitemOpen
  \bibfield  {author} {\bibinfo {author} {\bibfnamefont {Y.}~\bibnamefont {Fan}}, \bibinfo {author} {\bibfnamefont {G.}~\bibnamefont {Wen}}, \bibinfo {author} {\bibfnamefont {D.}~\bibnamefont {Li}}, \bibinfo {author} {\bibfnamefont {S.}~\bibnamefont {Qiu}}, \bibinfo {author} {\bibfnamefont {M.~D.}\ \bibnamefont {Levine}}, \ and\ \bibinfo {author} {\bibfnamefont {F.}~\bibnamefont {Xiao}},\ }\href {\doibase 10.1016/j.cviu.2020.102920} {\bibfield  {journal} {\bibinfo  {journal} {Computer Vision and Image Understanding}\ }\textbf {\bibinfo {volume} {195}},\ \bibinfo {pages} {102920} (\bibinfo {year} {2020})}\BibitemShut {NoStop}%
\bibitem [{\citenamefont {Yamaguchi}\ \emph {et~al.}(2021)\citenamefont {Yamaguchi}, \citenamefont {Hashimoto}, \citenamefont {Sugihara}, \citenamefont {Miyata}, \citenamefont {Murai}, \citenamefont {Takahashi}, \citenamefont {Honda}, \citenamefont {Hishimoto},\ and\ \citenamefont {Yamashita}}]{Yamaguchi2021}%
  \BibitemOpen
  \bibfield  {author} {\bibinfo {author} {\bibfnamefont {H.}~\bibnamefont {Yamaguchi}}, \bibinfo {author} {\bibfnamefont {Y.}~\bibnamefont {Hashimoto}}, \bibinfo {author} {\bibfnamefont {G.}~\bibnamefont {Sugihara}}, \bibinfo {author} {\bibfnamefont {J.}~\bibnamefont {Miyata}}, \bibinfo {author} {\bibfnamefont {T.}~\bibnamefont {Murai}}, \bibinfo {author} {\bibfnamefont {H.}~\bibnamefont {Takahashi}}, \bibinfo {author} {\bibfnamefont {M.}~\bibnamefont {Honda}}, \bibinfo {author} {\bibfnamefont {A.}~\bibnamefont {Hishimoto}}, \ and\ \bibinfo {author} {\bibfnamefont {Y.}~\bibnamefont {Yamashita}},\ }\href {\doibase 10.3389/fnins.2021.652987} {\bibfield  {journal} {\bibinfo  {journal} {Frontiers in Neuroscience}\ }\textbf {\bibinfo {volume} {15}} (\bibinfo {year} {2021}),\ 10.3389/fnins.2021.652987}\BibitemShut {NoStop}%
\bibitem [{\citenamefont {Prifti}\ \emph {et~al.}(2023)\citenamefont {Prifti}, \citenamefont {Buban}, \citenamefont {Thind},\ and\ \citenamefont {Klie}}]{Prifti2023}%
  \BibitemOpen
  \bibfield  {author} {\bibinfo {author} {\bibfnamefont {E.}~\bibnamefont {Prifti}}, \bibinfo {author} {\bibfnamefont {J.~P.}\ \bibnamefont {Buban}}, \bibinfo {author} {\bibfnamefont {A.~S.}\ \bibnamefont {Thind}}, \ and\ \bibinfo {author} {\bibfnamefont {R.~F.}\ \bibnamefont {Klie}},\ }\href {\doibase 10.1002/smll.202205977} {\bibfield  {journal} {\bibinfo  {journal} {Small}\ }\textbf {\bibinfo {volume} {19}} (\bibinfo {year} {2023}),\ 10.1002/smll.202205977}\BibitemShut {NoStop}%
\bibitem [{\citenamefont {Ayyubi}\ \emph {et~al.}(2024)\citenamefont {Ayyubi}, \citenamefont {Buban},\ and\ \citenamefont {Klie}}]{Ayyubi2024}%
  \BibitemOpen
  \bibfield  {author} {\bibinfo {author} {\bibfnamefont {R.~A.~W.}\ \bibnamefont {Ayyubi}}, \bibinfo {author} {\bibfnamefont {J.~P.}\ \bibnamefont {Buban}}, \ and\ \bibinfo {author} {\bibfnamefont {R.~F.}\ \bibnamefont {Klie}},\ }\href {\doibase 10.1093/mam/ozae044.180} {\bibfield  {journal} {\bibinfo  {journal} {Microscopy and Microanalysis}\ }\textbf {\bibinfo {volume} {30}} (\bibinfo {year} {2024}),\ 10.1093/mam/ozae044.180}\BibitemShut {NoStop}%
\bibitem [{\citenamefont {Lecun}\ \emph {et~al.}(1998)\citenamefont {Lecun}, \citenamefont {Bottou}, \citenamefont {Bengio},\ and\ \citenamefont {Haffner}}]{Lecun1998}%
  \BibitemOpen
  \bibfield  {author} {\bibinfo {author} {\bibfnamefont {Y.}~\bibnamefont {Lecun}}, \bibinfo {author} {\bibfnamefont {L.}~\bibnamefont {Bottou}}, \bibinfo {author} {\bibfnamefont {Y.}~\bibnamefont {Bengio}}, \ and\ \bibinfo {author} {\bibfnamefont {P.}~\bibnamefont {Haffner}},\ }\href {\doibase 10.1109/5.726791} {\bibfield  {journal} {\bibinfo  {journal} {Proceedings of the IEEE}\ }\textbf {\bibinfo {volume} {86}},\ \bibinfo {pages} {2278} (\bibinfo {year} {1998})}\BibitemShut {NoStop}%
\bibitem [{\citenamefont {Higgins}\ \emph {et~al.}(2017)\citenamefont {Higgins}, \citenamefont {Matthey}, \citenamefont {Pal}, \citenamefont {Burgess}, \citenamefont {Glorot}, \citenamefont {Botvinick}, \citenamefont {Mohamed},\ and\ \citenamefont {Lerchner}}]{Higgins2017}%
  \BibitemOpen
  \bibfield  {author} {\bibinfo {author} {\bibfnamefont {I.}~\bibnamefont {Higgins}}, \bibinfo {author} {\bibfnamefont {L.}~\bibnamefont {Matthey}}, \bibinfo {author} {\bibfnamefont {A.}~\bibnamefont {Pal}}, \bibinfo {author} {\bibfnamefont {C.}~\bibnamefont {Burgess}}, \bibinfo {author} {\bibfnamefont {X.}~\bibnamefont {Glorot}}, \bibinfo {author} {\bibfnamefont {M.}~\bibnamefont {Botvinick}}, \bibinfo {author} {\bibfnamefont {S.}~\bibnamefont {Mohamed}}, \ and\ \bibinfo {author} {\bibfnamefont {A.}~\bibnamefont {Lerchner}},\ }in\ \href {https://openreview.net/forum?id=Sy2fzU9gl} {\emph {\bibinfo {booktitle} {International Conference on Learning Representations}}}\ (\bibinfo {year} {2017})\BibitemShut {NoStop}%
\bibitem [{\citenamefont {Pearson}\ and\ \citenamefont {Galton}(1895)}]{Pearson1895}%
  \BibitemOpen
  \bibfield  {author} {\bibinfo {author} {\bibfnamefont {K.}~\bibnamefont {Pearson}}\ and\ \bibinfo {author} {\bibfnamefont {F.}~\bibnamefont {Galton}},\ }\href {\doibase 10.1098/rspl.1895.0041} {\bibfield  {journal} {\bibinfo  {journal} {Proceedings of the Royal Society of London}\ }\textbf {\bibinfo {volume} {58}},\ \bibinfo {pages} {240} (\bibinfo {year} {1895})}\BibitemShut {NoStop}%
\bibitem [{\citenamefont {Otsu}(1979)}]{Otsu1979}%
  \BibitemOpen
  \bibfield  {author} {\bibinfo {author} {\bibfnamefont {N.}~\bibnamefont {Otsu}},\ }\href {\doibase 10.1109/TSMC.1979.4310076} {\bibfield  {journal} {\bibinfo  {journal} {IEEE Transactions on Systems, Man, and Cybernetics}\ }\textbf {\bibinfo {volume} {9}},\ \bibinfo {pages} {62} (\bibinfo {year} {1979})}\BibitemShut {NoStop}%
\bibitem [{\citenamefont {Van~Rijsbergen}(1979)}]{Van1979}%
  \BibitemOpen
  \bibfield  {author} {\bibinfo {author} {\bibfnamefont {C.}~\bibnamefont {Van~Rijsbergen}},\ }in\ \href@noop {} {\emph {\bibinfo {booktitle} {Proceedings of the joint IBM/University of Newcastle upon tyne seminar on data base systems}}},\ Vol.~\bibinfo {volume} {79}\ (\bibinfo {year} {1979})\ pp.\ \bibinfo {pages} {1--14}\BibitemShut {NoStop}%
\bibitem [{\citenamefont {Rombach}\ \emph {et~al.}(2022)\citenamefont {Rombach}, \citenamefont {Blattmann}, \citenamefont {Lorenz}, \citenamefont {Esser},\ and\ \citenamefont {Ommer}}]{rombach2022}%
  \BibitemOpen
  \bibfield  {author} {\bibinfo {author} {\bibfnamefont {R.}~\bibnamefont {Rombach}}, \bibinfo {author} {\bibfnamefont {A.}~\bibnamefont {Blattmann}}, \bibinfo {author} {\bibfnamefont {D.}~\bibnamefont {Lorenz}}, \bibinfo {author} {\bibfnamefont {P.}~\bibnamefont {Esser}}, \ and\ \bibinfo {author} {\bibfnamefont {B.}~\bibnamefont {Ommer}},\ }\href {https://arxiv.org/abs/2112.10752} {\enquote {\bibinfo {title} {High-resolution image synthesis with latent diffusion models},}\ } (\bibinfo {year} {2022}),\ \Eprint {http://arxiv.org/abs/2112.10752}{arXiv:2112.10752 [cs.CV]}\BibitemShut {NoStop}%
\bibitem [{\citenamefont {Biswas}\ \emph {et~al.}(2023)\citenamefont {Biswas}, \citenamefont {Ziatdinov},\ and\ \citenamefont {Kalinin}}]{Biswas2023}%
  \BibitemOpen
  \bibfield  {author} {\bibinfo {author} {\bibfnamefont {A.}~\bibnamefont {Biswas}}, \bibinfo {author} {\bibfnamefont {M.}~\bibnamefont {Ziatdinov}}, \ and\ \bibinfo {author} {\bibfnamefont {S.~V.}\ \bibnamefont {Kalinin}},\ }\href {\doibase 10.1088/2632-2153/acf6a9} {\bibfield  {journal} {\bibinfo  {journal} {Machine Learning: Science and Technology}\ }\textbf {\bibinfo {volume} {4}},\ \bibinfo {pages} {045004} (\bibinfo {year} {2023})}\BibitemShut {NoStop}%
\bibitem [{\citenamefont {Sun}\ \emph {et~al.}(2018)\citenamefont {Sun}, \citenamefont {Wang}, \citenamefont {Xiong},\ and\ \citenamefont {Shao}}]{Sun2018}%
  \BibitemOpen
  \bibfield  {author} {\bibinfo {author} {\bibfnamefont {J.}~\bibnamefont {Sun}}, \bibinfo {author} {\bibfnamefont {X.}~\bibnamefont {Wang}}, \bibinfo {author} {\bibfnamefont {N.}~\bibnamefont {Xiong}}, \ and\ \bibinfo {author} {\bibfnamefont {J.}~\bibnamefont {Shao}},\ }\href {\doibase 10.1109/ACCESS.2018.2848210} {\bibfield  {journal} {\bibinfo  {journal} {IEEE Access}\ }\textbf {\bibinfo {volume} {6}},\ \bibinfo {pages} {33353} (\bibinfo {year} {2018})}\BibitemShut {NoStop}%
\bibitem [{\citenamefont {Paszke}\ \emph {et~al.}(2019)\citenamefont {Paszke}, \citenamefont {Gross}, \citenamefont {Massa}, \citenamefont {Lerer}, \citenamefont {Bradbury}, \citenamefont {Chanan}, \citenamefont {Killeen}, \citenamefont {Lin}, \citenamefont {Gimelshein}, \citenamefont {Antiga}, \citenamefont {Desmaison}, \citenamefont {Köpf}, \citenamefont {Yang}, \citenamefont {DeVito}, \citenamefont {Raison}, \citenamefont {Tejani}, \citenamefont {Chilamkurthy}, \citenamefont {Steiner}, \citenamefont {Fang}, \citenamefont {Bai},\ and\ \citenamefont {Chintala}}]{Paszke2019}%
  \BibitemOpen
  \bibfield  {author} {\bibinfo {author} {\bibfnamefont {A.}~\bibnamefont {Paszke}}, \bibinfo {author} {\bibfnamefont {S.}~\bibnamefont {Gross}}, \bibinfo {author} {\bibfnamefont {F.}~\bibnamefont {Massa}}, \bibinfo {author} {\bibfnamefont {A.}~\bibnamefont {Lerer}}, \bibinfo {author} {\bibfnamefont {J.}~\bibnamefont {Bradbury}}, \bibinfo {author} {\bibfnamefont {G.}~\bibnamefont {Chanan}}, \bibinfo {author} {\bibfnamefont {T.}~\bibnamefont {Killeen}}, \bibinfo {author} {\bibfnamefont {Z.}~\bibnamefont {Lin}}, \bibinfo {author} {\bibfnamefont {N.}~\bibnamefont {Gimelshein}}, \bibinfo {author} {\bibfnamefont {L.}~\bibnamefont {Antiga}}, \bibinfo {author} {\bibfnamefont {A.}~\bibnamefont {Desmaison}}, \bibinfo {author} {\bibfnamefont {A.}~\bibnamefont {Köpf}}, \bibinfo {author} {\bibfnamefont {E.}~\bibnamefont {Yang}}, \bibinfo {author} {\bibfnamefont {Z.}~\bibnamefont {DeVito}}, \bibinfo {author} {\bibfnamefont {M.}~\bibnamefont {Raison}}, \bibinfo {author} {\bibfnamefont {A.}~\bibnamefont {Tejani}}, \bibinfo
  {author} {\bibfnamefont {S.}~\bibnamefont {Chilamkurthy}}, \bibinfo {author} {\bibfnamefont {B.}~\bibnamefont {Steiner}}, \bibinfo {author} {\bibfnamefont {L.}~\bibnamefont {Fang}}, \bibinfo {author} {\bibfnamefont {J.}~\bibnamefont {Bai}}, \ and\ \bibinfo {author} {\bibfnamefont {S.}~\bibnamefont {Chintala}},\ }\href {https://arxiv.org/abs/1912.01703} {\enquote {\bibinfo {title} {Pytorch: An imperative style, high-performance deep learning library},}\ } (\bibinfo {year} {2019}),\ \Eprint {http://arxiv.org/abs/1912.01703}{arXiv:1912.01703 [cs.LG]}\BibitemShut {NoStop}%
\bibitem [{\citenamefont {Pennycook}\ and\ \citenamefont {Nellist}(2011)}]{STEM2011}%
  \BibitemOpen
  \bibinfo {editor} {\bibfnamefont {S.~J.}\ \bibnamefont {Pennycook}}\ and\ \bibinfo {editor} {\bibfnamefont {P.~D.}\ \bibnamefont {Nellist}},\ eds.,\ \href {\doibase 10.1007/978-1-4419-7200-2} {\emph {\bibinfo {title} {Scanning Transmission Electron Microscopy}}}\ (\bibinfo  {publisher} {Springer New York},\ \bibinfo {year} {2011})\BibitemShut {NoStop}%
\bibitem [{\citenamefont {An}\ and\ \citenamefont {Cho}(2015{\natexlab{a}})}]{An2015}%
  \BibitemOpen
  \bibfield  {author} {\bibinfo {author} {\bibfnamefont {J.}~\bibnamefont {An}}\ and\ \bibinfo {author} {\bibfnamefont {S.}~\bibnamefont {Cho}},\ }\href {https://api.semanticscholar.org/CorpusID:36663713} {\enquote {\bibinfo {title} {Variational autoencoder based anomaly detection using reconstruction probability},}\ } (\bibinfo {year} {2015}{\natexlab{a}})\BibitemShut {NoStop}%
\bibitem [{\citenamefont {Matsuo}\ \emph {et~al.}(2017)\citenamefont {Matsuo}, \citenamefont {Fukuhara},\ and\ \citenamefont {Shimada}}]{Matsuo2017}%
  \BibitemOpen
  \bibfield  {author} {\bibinfo {author} {\bibfnamefont {T.}~\bibnamefont {Matsuo}}, \bibinfo {author} {\bibfnamefont {H.}~\bibnamefont {Fukuhara}}, \ and\ \bibinfo {author} {\bibfnamefont {N.}~\bibnamefont {Shimada}},\ }\href {https://arxiv.org/abs/1709.03754} {\enquote {\bibinfo {title} {Transform invariant auto-encoder},}\ } (\bibinfo {year} {2017}),\ \Eprint {http://arxiv.org/abs/1709.03754}{arXiv:1709.03754 [cs.CV]}\BibitemShut {NoStop}%
\bibitem [{\citenamefont {Ng}\ and\ \citenamefont {Yang}(2023)}]{Ng2023}%
  \BibitemOpen
  \bibfield  {author} {\bibinfo {author} {\bibfnamefont {K.-K.}\ \bibnamefont {Ng}}\ and\ \bibinfo {author} {\bibfnamefont {M.-F.}\ \bibnamefont {Yang}},\ }\href {\doibase 10.1103/PhysRevB.108.214428} {\bibfield  {journal} {\bibinfo  {journal} {Physical Review B}\ }\textbf {\bibinfo {volume} {108}},\ \bibinfo {pages} {214428} (\bibinfo {year} {2023})}\BibitemShut {NoStop}%
\bibitem [{\citenamefont {Cheng}\ \emph {et~al.}(2021)\citenamefont {Cheng}, \citenamefont {Zhu}, \citenamefont {Wang}, \citenamefont {Zhang},\ and\ \citenamefont {Li}}]{Cheng2021}%
  \BibitemOpen
  \bibfield  {author} {\bibinfo {author} {\bibfnamefont {Z.}~\bibnamefont {Cheng}}, \bibinfo {author} {\bibfnamefont {E.}~\bibnamefont {Zhu}}, \bibinfo {author} {\bibfnamefont {S.}~\bibnamefont {Wang}}, \bibinfo {author} {\bibfnamefont {P.}~\bibnamefont {Zhang}}, \ and\ \bibinfo {author} {\bibfnamefont {W.}~\bibnamefont {Li}},\ }\href {\doibase 10.1109/ACCESS.2021.3065838} {\bibfield  {journal} {\bibinfo  {journal} {IEEE Access}\ }\textbf {\bibinfo {volume} {9}},\ \bibinfo {pages} {43991} (\bibinfo {year} {2021})}\BibitemShut {NoStop}%
\bibitem [{\citenamefont {An}\ and\ \citenamefont {Cho}(2015{\natexlab{b}})}]{Jinwon2015}%
  \BibitemOpen
  \bibfield  {author} {\bibinfo {author} {\bibfnamefont {J.}~\bibnamefont {An}}\ and\ \bibinfo {author} {\bibfnamefont {S.}~\bibnamefont {Cho}},\ }\href@noop {} {\enquote {\bibinfo {title} {Variational autoencoder based anomaly detection using reconstruction probability},}\ } (\bibinfo {year} {2015}{\natexlab{b}})\BibitemShut {NoStop}%
\end{thebibliography}%

\onecolumngrid  
\clearpage  
\section{End Matter}

\begin{figure}[h!]  
    \includegraphics[width=1\textwidth]{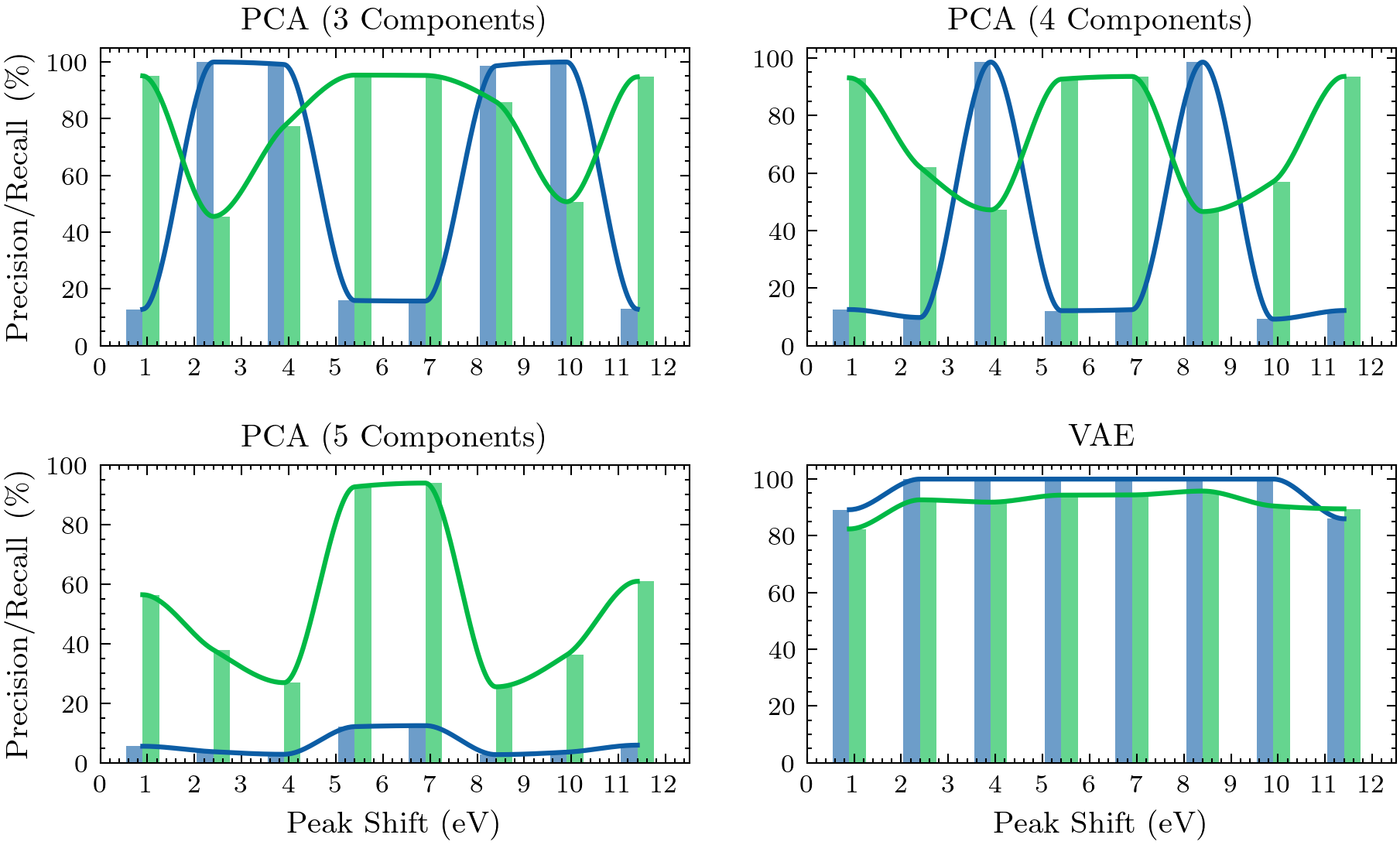}
    \caption{Precision-recall curves comparing anomaly detection performance between our 3D-CVAE model and PCA with 3,4,5 components. The VAE demonstrates consistently superior performance across different detection thresholds, maintaining a better balance between precision and recall. In contrast, PCA exhibits trade-off behavior, where improved precision comes at the cost of substantially reduced recall and vice versa, indicating less reliable anomaly detection.}
    \label{fig:precision_recall_plot}
\end{figure}

\begin{figure}[h!]  
    \includegraphics[width=1\textwidth]{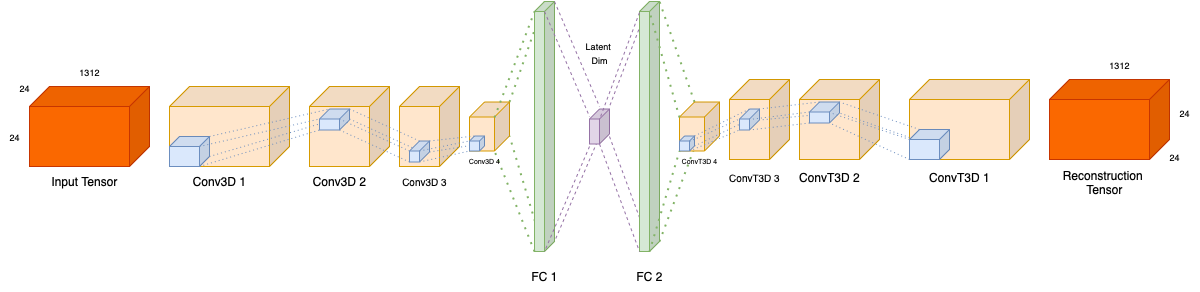}
    \caption{Schematic representation of the 3D-CVAE architecture. The encoder processes the input EELS SI datacube through convolutional layers to a compressed latent space representation. The decoder then reconstructs the data from this latent space back to the original dimensionality.}
    \label{fig:CVAE_Schematic}
\end{figure}

\end{document}